\pdfoutput=1

\documentclass[11pt]{article}

\usepackage[final]{coling}

\usepackage{times}
\usepackage{latexsym}

\usepackage[T1]{fontenc}
\usepackage{tabularx} 
\usepackage{graphicx}
\usepackage{multirow}
\usepackage[utf8]{inputenc}

\usepackage{microtype}

\usepackage{inconsolata}

\usepackage{graphicx}
\usepackage{amsmath}

\title{LuxVeri at GenAI Detection Task 1: Inverse Perplexity Weighted Ensemble for Robust Detection of AI-Generated Text across English and Multilingual Contexts}

\author{
 \textbf{Md Kamrujjaman Mobin\textsuperscript{1}},
 \textbf{Md Saiful Islam\textsuperscript{1, 2}} 
\\
 \textsuperscript{1}Computer Science and Engineering, Shahjalal University of Science and Technology,\\ Sylhet, Bangladesh\\
 \textsuperscript{2}Computing Science,  University of Alberta, Edmonton, Alberta, Canada\\
}

\begin{document}
\maketitle
\begin{abstract}
This paper presents a system developed for Task 1 of the COLING 2025 Workshop on Detecting AI-Generated Content, focusing on the binary classification of machine-generated versus human-written text. Our approach utilizes an ensemble of models, with weights assigned according to each model's inverse perplexity, to enhance classification accuracy. For the English text detection task, we combined RoBERTa-base, RoBERTa-base with the OpenAI detector, and BERT-base-cased, achieving a Macro F1-score of 0.7458, which ranked us 12th out of 35 teams. We ensembled RemBERT, XLM-RoBERTa-base, and BERT-base-multilingual-case for the multilingual text detection task, employing the same inverse perplexity weighting technique. This resulted in a Macro F1-score of 0.7513, positioning us 4th out of 25 teams. Our results demonstrate the effectiveness of inverse perplexity weighting in improving the robustness of machine-generated text detection across both monolingual and multilingual settings, highlighting the potential of ensemble methods for this challenging task.
\end{abstract}

\section{Introduction}

The rapid advancement of language models such as GPT \cite{radford2019language} and BERT \cite{devlin2019bert} has increased machine-generated content, raising significant concerns about misinformation and academic integrity. Identifying AI-generated text becomes more challenging in multilingual contexts, where linguistic diversity adds further complexity to model generalization. While existing approaches perform well in English, their effectiveness decreases when applied to languages with diverse syntactic and semantic structures.

In Task 1 of the COLING 2025 Workshop on Detecting AI-Generated Content \cite{wang2025genai}, we propose an ensemble-based solution to address these issues. For English detection, we combine RoBERTa-base \cite{liu2019roberta}, OpenAI’s AI text detector \cite{solaiman2019release}, and BERT-base-cased \cite{devlin2019bert}. For multilingual detection, we integrate RemBERT \cite{DBLP:conf/iclr/ChungFTJR21}, XLM-RoBERTa-base \cite{DBLP:journals/corr/abs-1911-02116}, and BERT-base-multilingual-cased \cite{devlin2019bert}. To further improve performance, we incorporate inverse perplexity weighting to give greater priority to models that produce lower perplexity scores. Our ensemble approach achieved a Macro F1-score of 0.7458 (Micro F1: 0.7568) in English, placing us 12th out of 35 teams, and a Macro F1-score of 0.7513 (Micro F1: 0.7527) for the multilingual tasks, ranking 4th out of 25 teams.

We encountered several challenges during this work. One major issue was data imbalance, as human-written content vastly outnumbered AI-generated samples. To address this, we employed data augmentation and optimized our sampling strategies. Another challenge involved ensuring the models’ generalization across different languages and writing styles, often with limited training data. This highlights the importance of additional fine-tuning and the need to explore alternative architectures that can better handle diverse linguistic inputs.

This paper presents a robust ensemble approach for detecting AI-generated content, with strong performance across both English and multilingual tasks. However, significant opportunities remain for improving model generalization and addressing data imbalance, which will be crucial for future advancements in this field. The following sections will discuss the dataset, methodology, results, a detailed analysis of the findings, and conclusions drawn from this study.


\begin{table*}[ht]
    \centering
    \small
    \renewcommand{\arraystretch}{1.1} 
    \setlength{\tabcolsep}{1.5pt} 
    \begin{tabular}{p{1.5cm}|>{\centering\arraybackslash}p{1.5cm}|>{\centering\arraybackslash}p{1.5cm}|>{\centering\arraybackslash}p{1.5cm}|>{\centering\arraybackslash}p{1.5cm}|>{\centering\arraybackslash}p{1.5cm}|>{\centering\arraybackslash}p{1.5cm}|>{\centering\arraybackslash}p{1.5cm}|>{\centering\arraybackslash}p{1.5cm}|>{\centering\arraybackslash}p{1.5cm}}
        \hline
        \hline
        \textbf{Dataset} & \textbf{en} & \textbf{zh} & \textbf{bg} & \textbf{de} & \textbf{it} & \textbf{id} & \textbf{ur} & \textbf{ar} & \textbf{ru} \\
        \hline
        \textbf{Training} & 610,676 & 35,284 & 8,091 & 4,693 & 4,174 & 3,976 & 3,761 & 2,114 & 1,314 \\
        \hline
        \textbf{Validation} & 261,849 & 14,772 & 3,489 & 2,059 & 1,843 & 1,803 & 1,573 & 906 & 600 \\
        \hline
    \end{tabular}
    \caption{Data distribution for multilingual training and validation datasets, including the number of entries per language. The English dataset is consistent across both English-only and multilingual contexts, so it is omitted from the table for clarity.}
    \label{tab:data_distribution}
\end{table*}

\section{Background}
\subsection{Dataset}
The provided dataset includes training and validation sets for two subtasks: Subtask A (English-only Machine-Generated Text Detection) and Subtask B (Multilingual Machine-Generated Text Detection). Subtask A consists of over 610,000 English-only entries in the training set and around 261,000 in the validation set, each labeled as machine-generated or human-generated. These texts are sourced from various platforms, with information on their origin and creation model (e.g., GPT-4, human). Subtask B extends the dataset to include over 674,000 training entries and approximately 288,000 validation entries across nine languages, including English, Chinese, and Bulgarian. Each entry contains details on the source, sub-source, language, model, label, and text. Data distribution details are shown in Table \ref{tab:data_distribution}.

\subsection{Related Work}

The detection of AI-generated text has garnered significant attention with the advent of large language models (LLMs) such as GPT \cite{radford2019language} and BERT \cite{devlin2019bert}. Fine-tuning Transformer-based models for binary classification has shown efficacy; however, challenges persist, particularly in multilingual settings where data biases impede generalization \cite{zellers2019defending, solaiman2019release}.  

Ensemble methods combining BERT, RoBERTa, and GPT variants have enhanced robustness across domains and languages \cite{schick2020exploiting}. Perplexity-based weighting strategies further optimize individual model contributions \cite{clark2019ensemble}. Multilingual models like XLM-RoBERTa \cite{DBLP:journals/corr/abs-1911-02116} and RemBERT \cite{DBLP:conf/iclr/ChungFTJR21} improve cross-lingual performance, though low-resource languages remain challenging \cite{hu2020multilingual}.

Recent advancements in shared tasks, such as those introduced by SemEval \cite{fetahu-etal-2023-semeval, wang2024semeval2024task8multidomain}, have refined methodologies through task-specific fine-tuning and the integration of multilingual pre-trained models \cite{eger2023semeval, siino-2024-badrock}.  

Building upon these foundations, our work employs an inverse perplexity-weighted ensemble approach to optimize model contributions, enhancing robustness in both monolingual and multilingual detection scenarios.

\begin{figure}[ht]
    \centering
    \includegraphics[width=0.8\columnwidth]{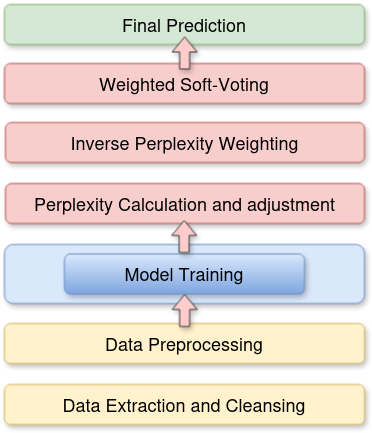}
    \caption{Overall Framework of our Proposed System.}
    \label{fig:system_overview}
\end{figure}

\begin{table*}[ht]
    \centering
    \small
    \renewcommand{\arraystretch}{1.1} 
    \setlength{\tabcolsep}{1.5pt} 
    \begin{tabular}{p{1.5cm}|>{\centering\arraybackslash}p{1.5cm}|>{\centering\arraybackslash}p{1.5cm}|>{\centering\arraybackslash}p{1.5cm}|>{\centering\arraybackslash}p{1.5cm}|>{\centering\arraybackslash}p{1.5cm}|>{\centering\arraybackslash}p{1.5cm}|>{\centering\arraybackslash}p{1.5cm}|>{\centering\arraybackslash}p{1.5cm}|>{\centering\arraybackslash}p{1.5cm}}
        \hline
        \hline
        \textbf{Dataset} & \textbf{en} & \textbf{zh} & \textbf{bg} & \textbf{de} & \textbf{it} & \textbf{id} & \textbf{ur} & \textbf{ar} & \textbf{ru} \\
        \hline
        \textbf{Training} & 40,000 & 20,000 & 8,091 & 4,693 & 4,174 & 3,976 & 3,761 & 2,114 & 1,314 \\
        \hline
        \textbf{Validation} & 26,000 & 10,000 & 3,489 & 2,059 & 1,843 & 1,803 & 1,573 & 906 & 600 \\
        \hline
    \end{tabular}
    \caption{Reduced data distribution for the multilingual task after balancing by scaling down English and Chinese samples to improve performance in underrepresented languages.}
    \label{tab:data_distribution_reduced}
\end{table*}

\section{System Overview}
We developed an ensemble approach for AI-generated text detection across English and multilingual contexts, using Transformer models with a weighted voting strategy based on inverse perplexity for improved accuracy. The system overview is shown in Figure \ref{fig:system_overview}.

\subsection{Ensemble Model Selection and Justification}
We selected six Transformer-based models for our ensemble: three for English and three for multilingual contexts, chosen for their ability to capture linguistic and syntactic patterns.

\begin{itemize}
    \item \textbf{English Models}: The models utilized in our work include \textbf{RoBERTa-base}, renowned for its robust performance in natural language understanding, effectively capturing deep syntactic and semantic patterns \cite{liu2019roberta}. Additionally, the \textbf{RoBERTa-base OpenAI detector} is fine-tuned to detect AI-generated content by identifying subtle machine-generated patterns \cite{solaiman2019release}. Lastly, \textbf{BERT-base-cased} is incorporated for its capability to handle case-sensitive distinctions, which are critical for nuanced classification tasks \cite{devlin2019bert}.
    \item \textbf{Multilingual Models}: For Multilingual, we employed \textbf{RemBERT}, a model optimized for multilingual tasks, demonstrating exceptional performance in syntactic and semantic understanding across languages \cite{DBLP:conf/iclr/ChungFTJR21}. Furthermore, \textbf{XLM-RoBERTa-base} is employed for its strength in cross-lingual applications, adeptly handling diverse language structures \cite{DBLP:journals/corr/abs-1911-02116}. Additionally, \textbf{BERT-base-multilingual-cased} is used as it is specifically designed to capture linguistic diversity and perform effectively in multilingual tasks \cite{devlin2019bert}.
\end{itemize}

We trained these models on the dataset provided as part of the shared task 1 \cite{wang2025genai}, including human-authored and AI-generated content. This enabled the ensemble to generalize effectively across both English and multilingual contexts.

\subsection{Data Pre-processing}

The multilingual task presented a significant data imbalance across languages, as shown in Table \ref{tab:data_distribution}, which details the original distribution of training and validation samples. For instance, the dataset included 610,676 English (en) samples and 35,284 Chinese (zh) samples, whereas underrepresented languages like Urdu (ur), Arabic (ar), and Russian (ru) had far fewer samples (3,761, 2,114, and 1,314, respectively). This imbalance hindered the model's ability to predict outputs for these underrepresented languages accurately.   

To mitigate this issue, we implemented a dataset balancing strategy by reducing the number of English and Chinese samples to a proportionate scale. This adjustment enabled the model to better focus on learning patterns in the underrepresented languages, thereby enhancing overall performance and reducing biases in predictions. The detailed data distribution after applying this balancing strategy is presented in Table \ref{tab:data_distribution_reduced}.  

Following this, text data was processed using model-specific tokenizers, with truncation and padding applied as needed. To optimize memory usage and training efficiency, text length was calculated and sorted by word count, minimizing unnecessary padding. A fixed random seed was used throughout to ensure reproducibility.

\begin{table}[ht]
\centering
\small
\renewcommand{\arraystretch}{1.3} 
\setlength{\tabcolsep}{3pt} 
\begin{tabular}{>{\raggedright\arraybackslash}p{3.5cm}|>{\raggedright\arraybackslash}p{3.5cm}} 
\hline \hline
\textbf{Hyperparameter} & \textbf{Value} \\ \hline 
Number of Epochs        & $2\sim3$ \\ 
Learning Rate           & $1 \times 10^{-5} \sim 2 \times 10^{-5}$ \\ 
Training Batch Size     & 4 \\ 
Validation Batch Size   & 16 \\ 
Early Stopping Patience & 5 validation steps \\ 
Early Stopping Threshold& 0.001 \\ 
Weight Decay            & 0.01 \\ 
Optimizer               & AdamW \\ 
Loss Function           & Binary Cross-Entropy \\ 
Evaluation Strategy     & Every ¼ epoch \\ 
Checkpointing Strategy  & Validation loss \\ 
\end{tabular}
\caption{Training Configuration}
\label{table:training_config}
\end{table}

\begin{table*}[ht]
    \centering
    \small
    \renewcommand{\arraystretch}{1.45} 
    \setlength{\tabcolsep}{8pt} 
    \begin{tabular}{>{\centering\arraybackslash}p{3.5cm}|>{\centering\arraybackslash}p{0.8cm}|>{\centering\arraybackslash}p{0.8cm}|>{\centering\arraybackslash}p{0.7cm}||>{\centering\arraybackslash}p{3cm}|>{\centering\arraybackslash}p{0.8cm}|>{\centering\arraybackslash}p{0.8cm}|>{\centering\arraybackslash}p{0.5cm}}
        \hline \hline
        \multicolumn{4}{c||}{\textbf{English-only Task}} & \multicolumn{4}{c}{\textbf{Multilingual Task}} \\
        \hline 
        \textbf{Model} & \textbf{Micro F1-Score} & \textbf{Macro F1-Score} & \textbf{Rank} & \textbf{Model} & \textbf{Micro F1-Score} & \textbf{Macro F1-Score} & \textbf{Rank} \\
        \hline \hline
        RoBERTa + RoBERTa OpenAI + BERT cased & 0.7568 & 0.7458 & 12/35 & RemBERT + XLM-R + BERT multilingual & 0.7527 & 0.7513 & 4/25 \\
        \hline
        RoBERTa + RoBERTa OpenAI & 0.7495 & 0.7380 & -- & RemBERT & 0.7507 & 0.7489 & -- \\
        \hline
        Baseline(RoBERTa)* & 0.7381 & 0.7342 & -- & RemBERT + XLM-R & 0.7473 & 0.7435 & -- \\
        \hline
        RoBERTa + BERT cased & 0.7275 & 0.7229 & -- & Baseline(XLM-R)* & 0.7426 & 0.7416 & -- \\
        \hline
    \end{tabular}
    \caption{Performance of various experimented models on English-only and multilingual tasks compared with baseline results.}
    \label{tab:performance_results}
\end{table*}

\subsection{Training Procedure}  

The model was fine-tuned using the Hugging Face Transformers library\footnote{Hugging Face Transformers: \url{https://huggingface.co/transformers/}} for both English and multilingual text classification tasks. Datasets were processed into Hugging Face Dataset objects, with tokenization performed using AutoTokenizer for models like RemBERT and RoBERTa-base. The architecture was adapted for classification tasks with appropriate label mappings.

Key hyperparameters, including learning rate, batch size, and weight decay, were optimized through empirical experiments to balance performance and efficiency. Learning rates between $1 \times 10^{-5}$ and $2 \times 10^{-5}$ were tested, with lower rates promoting smoother convergence. A batch size of 4 for training and 16 for validation balanced memory and efficiency.

Hyperparameter tuning reduced overfitting and improved generalization, leading to faster convergence and better validation performance. Early stopping (patience of 5 steps, threshold of 0.001) prevented overfitting and enhanced robustness.

Checkpoints were saved for each epoch, and the best model was retained for testing. Further details of the final configurations are in Table \ref{table:training_config}, ensuring effective fine-tuning across both datasets.

\subsection{Ensemble Voting Strategy}
Our ensemble employs a weighted soft-voting strategy, combining predictions from three different models for each subtask. The weights are determined based on inverse perplexity, with lower perplexity values reflecting higher confidence.

\subsubsection{Perplexity Calculation}
For each model, we compute the perplexity based on its predictions. The perplexity \( P \) is computed using the Negative Log Likelihood formula:

\[
P = \exp \left( - \frac{1}{N} \sum_{i=1}^{N} \log(p(y_i \mid x_i)) \right)
\]

where \( p(y_i \mid x_i) \) is the predicted probability for the true label \( y_i \), and \( N \) is the number of test samples. Lower perplexity values indicate higher confidence.

To compute perplexity, we use each model's logits, apply softmax to obtain probabilities, and then calculate perplexity based on the true labels and these probabilities.

\subsubsection{Perplexity-Based Weighting Adjustment}

To calculate model weights, each model's perplexity is adjusted by subtracting 1, creating an effective weighting scale. The weight \( w_i \) for model \( i \) is then computed as the inverse of this adjusted perplexity and normalized across models, giving higher confidence models greater influence.

\[
w_i = \frac{1 / (P_i - 1)}{\sum_{j=1}^{M} (1 / (P_j - 1))}
\]

where \( M \) represents the total number of models, and \( P_i \) is the original perplexity of model \( i \).

\subsubsection{Weighted Soft-Voting}
Each model's predicted probabilities are scaled by its weight and summed to form the final ensemble prediction. This weighted voting prioritizes models with higher confidence (lower perplexity), giving them greater influence on the final decision. The ensemble's final prediction for each class \( c \) is:

\[
p_{\text{ensemble}}(c) = \sum_{i=1}^{M} w_i \cdot p_i(c)
\]

where \( p_i(c) \) is the predicted probability for class \( c \) by model \( i \), and \( w_i \) is its weight.

This method enhances ensemble accuracy by prioritizing predictions from more confident models, improving overall performance.

\section{Results}

This section presents the performance of our ensemble approach for Task 1 at the COLING 2025 Workshop on Detecting AI-Generated Content, evaluated using the Macro F1-score. Detailed results are shown in Table \ref{tab:ensemble_comparison}.

The baseline scores provided by the task organizers \cite{wang2025genai} used RoBERTa-base \cite{liu2019roberta} for the English track and XLM-RoBERTa-base \cite{DBLP:journals/corr/abs-1911-02116} for the multilingual track. These scores serve as benchmarks for our ensemble method.

During testing, models in the ensemble were weighted based on perplexity, with lower-perplexity models given greater influence. This approach, along with sorting test data by text length, reduced inference time by 40\% while generating predictions using softmax and weighted averaging.

For the English-only task, our ensemble of RoBERTa-base, RoBERTa-base OpenAI detector, and BERT-base-cased achieved a Macro F1-score of 0.7458, outperforming the baseline (0.7342) and ranking 12th out of 35 teams. Similarly, for the multilingual task, our ensemble of RemBERT, XLM-RoBERTa-base, and BERT-base-multilingual-cased achieved a Macro F1-score of 0.7513, surpassing the baseline (0.7416) and ranking 4th out of 25 teams. The combination of these models effectively enhanced both language-specific and cross-lingual accuracy.

Table \ref{tab:ensemble_comparison} highlights the effectiveness of Inverse Perplexity Weighting, which achieved the highest Macro F1-scores for both tasks (0.7458 for English and 0.7513 for multilingual). This method dynamically prioritizes models with lower uncertainty in their predictions, outperforming other techniques such as accuracy-based weighting, mean ensembling, and majority voting.

Our ensemble also outperformed individual models, such as RoBERTa-base (0.7342) for the English track and the dual combination of RemBERT + XLM-R (0.7435) for the multilingual track. This demonstrates the effectiveness of combining diverse models to achieve better performance.

\begin{table}[ht]
    \centering
    \small
    \renewcommand{\arraystretch}{1.3} 
    \setlength{\tabcolsep}{5pt} 
    \begin{tabular}{>{\centering\arraybackslash}p{2.4cm}|>{\centering\arraybackslash}p{1.7cm}|>{\centering\arraybackslash}p{1.85cm}}
        \hline \hline
        \textbf{Ensemble Technique} & \textbf{English Task  (Macro F1)} & \textbf{Multilingual Task (Macro F1)} \\
        \hline \hline
        Inverse Perplexity Weighting & \textbf{0.7458} & \textbf{0.7513} \\
        \hline
        Accuracy Based Weighting & 0.7251 & 0.7393 \\
        \hline
        Mean Ensemble & 0.7153 & 0.7211 \\
        \hline
        Majority Voting & 0.6850 & 0.7005 \\
        \hline
    \end{tabular}
    \caption{Comparison of ensemble techniques for English-only and multilingual tasks, highlighting the effectiveness of Inverse Perplexity Weighting.}
    \label{tab:ensemble_comparison}
\end{table}

\section{Discussion and Conclusion}

In this work, we presented an ensemble approach to detect AI-generated content across English and multilingual datasets. By combining multiple pre-trained models, including RoBERTa-base, OpenAI detector, BERT-base-cased for English, and RemBERT, XLM-RoBERTa-base, BERT-base-multilingual-cased for multilingual tasks, and applying inverse perplexity weighting, our ensemble demonstrated strong performance. It achieved a Macro F1-score of 0.7458 (Micro F1: 0.7568) for English, ranking 12th, and 0.7513 (Micro F1: 0.7527) for multilingual tasks, ranking 4th.

Compared to individual models, our ensemble consistently outperformed or matched their performance. For example, in the English task, the ensemble scored 0.7568, surpassing RoBERTa + OpenAI detector (0.7381) and BERT (0.7275). Similarly, the multilingual task achieved 0.7527, exceeding RemBERT + XLM-R (0.7473) and RemBERT (0.7507). Notably, our ensemble also outperformed baseline models, with a Macro F1-score of 0.7458 for English (baseline: 0.7342 achieved by RoBERTa) and 0.7513 for multilingual (baseline: 0.7416 achieved by XLM-RoBERTa). These results highlight the effectiveness of combining model strengths to improve detection accuracy.

A key challenge faced during the multilingual task was the data imbalance between well-represented languages like English and Chinese and underrepresented ones such as Urdu, Arabic, and Russian. This disparity hindered the model's accuracy for underrepresented languages. To address this, we scale down samples from overrepresented languages to balance the dataset. This adjustment improved performance across languages, validating the effectiveness of our approach.

Despite these successes, challenges persist. Detecting AI-generated content in multilingual contexts remains complex and demands further refinement in model architectures and data processing techniques. Future work could explore advanced methods for mitigating data imbalance, such as data augmentation or active learning, to enhance the model's generalization ability across diverse languages. Additionally, more sophisticated ensemble strategies or domain-specific models could improve detection accuracy.

In conclusion, this study demonstrates the effectiveness of an ensemble approach for detecting AI-generated content across English and multilingual datasets. Addressing data imbalance and using inverse perplexity weighting improved performance, though ongoing challenges highlight the need for continuous innovation in AI detection systems.

\bibliography{coling_latex}

\begin{thebibliography}{15}
\providecommand{\natexlab}[1]{#1}

\bibitem[{Chung et~al.(2021)Chung, F{\'{e}}vry, Tsai, Johnson, and Ruder}]{DBLP:conf/iclr/ChungFTJR21}
Hyung~Won Chung, Thibault F{\'{e}}vry, Henry Tsai, Melvin Johnson, and Sebastian Ruder. 2021.
\newblock \href {https://openreview.net/forum?id=xpFFI\_NtgpW} {Rethinking embedding coupling in pre-trained language models}.
\newblock In \emph{9th International Conference on Learning Representations, {ICLR} 2021, Virtual Event, Austria, May 3-7, 2021}. OpenReview.net.

\bibitem[{Clark et~al.(2019)Clark, Gimpel, and Smith}]{clark2019ensemble}
Christopher Clark, Kevin Gimpel, and Noah~A Smith. 2019.
\newblock Ensemble methods for detection of machine-generated text.
\newblock In \emph{Proceedings of ACL 2019}, pages 47--56.

\bibitem[{Conneau et~al.(2019)Conneau, Khandelwal, Goyal, Chaudhary, Wenzek, Guzm{\'{a}}n, Grave, Ott, Zettlemoyer, and Stoyanov}]{DBLP:journals/corr/abs-1911-02116}
Alexis Conneau, Kartikay Khandelwal, Naman Goyal, Vishrav Chaudhary, Guillaume Wenzek, Francisco Guzm{\'{a}}n, Edouard Grave, Myle Ott, Luke Zettlemoyer, and Veselin Stoyanov. 2019.
\newblock \href {https://arxiv.org/abs/1911.02116} {Unsupervised cross-lingual representation learning at scale}.
\newblock \emph{CoRR}, abs/1911.02116.

\bibitem[{Devlin et~al.(2019)Devlin, Chang, Lee et~al.}]{devlin2019bert}
Jacob Devlin, Ming-Wei Chang, Kenton Lee, et~al. 2019.
\newblock \href {https://doi.org/10.18653/v1/N19-1423} {Bert: Pre-training of deep bidirectional transformers for language understanding}.
\newblock In \emph{Proceedings of NAACL-HLT 2019}, pages 4171--4186.

\bibitem[{Eger et~al.(2023)}]{eger2023semeval}
S.~Eger et~al. 2023.
\newblock Semeval-2023: Ai-generated text detection task.
\newblock In \emph{Proceedings of SemEval 2023}.

\bibitem[{Fetahu et~al.(2023)Fetahu, Kar, Chen, Rokhlenko, and Malmasi}]{fetahu-etal-2023-semeval}
Besnik Fetahu, Sudipta Kar, Zhiyu Chen, Oleg Rokhlenko, and Shervin Malmasi. 2023.
\newblock \href {https://doi.org/10.18653/v1/2023.semeval-1.310} {{S}em{E}val-2023 task 2: Fine-grained multilingual named entity recognition ({M}ulti{C}o{NER} 2)}.
\newblock In \emph{Proceedings of the 17th International Workshop on Semantic Evaluation (SemEval-2023)}, pages 2247--2265, Toronto, Canada. Association for Computational Linguistics.

\bibitem[{Hu et~al.(2020)Hu, Li, and Zhang}]{hu2020multilingual}
Jing Hu, Yizhe Li, and Zhiyuan Zhang. 2020.
\newblock Multilingual pretraining for cross-lingual natural language understanding.
\newblock In \emph{Proceedings of NAACL 2020}, pages 1--10.

\bibitem[{Liu et~al.(2019)}]{liu2019roberta}
Y.~Liu et~al. 2019.
\newblock \href {https://arxiv.org/abs/1907.11692} {Roberta: A robustly optimized bert pretraining approach}.
\newblock \emph{arXiv preprint arXiv:1907.11692}.

\bibitem[{Radford et~al.(2019)Radford, Wu, Child, Luan, Amodei, Sutskever et~al.}]{radford2019language}
Alec Radford, Jeffrey Wu, Rewon Child, David Luan, Dario Amodei, Ilya Sutskever, et~al. 2019.
\newblock Language models are unsupervised multitask learners.
\newblock \emph{OpenAI blog}, 1(8):9.

\bibitem[{Schick and Schütze(2020)}]{schick2020exploiting}
Timo Schick and Hinrich Schütze. 2020.
\newblock Exploiting cloze questions for few-shot text classification.
\newblock In \emph{Proceedings of ACL 2020}, pages 255--265.

\bibitem[{Siino(2024)}]{siino-2024-badrock}
Marco Siino. 2024.
\newblock \href {https://doi.org/10.18653/v1/2024.semeval-1.37} {{B}ad{R}ock at {S}em{E}val-2024 task 8: {D}istil{BERT} to detect multigenerator, multidomain and multilingual black-box machine-generated text}.
\newblock In \emph{Proceedings of the 18th International Workshop on Semantic Evaluation (SemEval-2024)}, pages 239--245, Mexico City, Mexico.

\bibitem[{Solaiman et~al.(2019)Solaiman, Brundage, Clark, Askell, Herbert-Voss, Wu, Radford, Krueger, Kim, Kreps et~al.}]{solaiman2019release}
Irene Solaiman, Miles Brundage, Jack Clark, Amanda Askell, Ariel Herbert-Voss, Jeff Wu, Alec Radford, Gretchen Krueger, Jong~Wook Kim, Sarah Kreps, et~al. 2019.
\newblock Release strategies and the social impacts of language models.
\newblock \emph{arXiv preprint arXiv:1908.09203}.

\bibitem[{Wang et~al.(2024)Wang, Mansurov, Ivanov, Su, Shelmanov, Tsvigun, Afzal, Mahmoud, Puccetti, Arnold, Whitehouse, Aji, Habash, Gurevych, and Nakov}]{wang2024semeval2024task8multidomain}
Yuxia Wang, Jonibek Mansurov, Petar Ivanov, Jinyan Su, Artem Shelmanov, Akim Tsvigun, Osama~Mohammed Afzal, Tarek Mahmoud, Giovanni Puccetti, Thomas Arnold, Chenxi Whitehouse, Alham~Fikri Aji, Nizar Habash, Iryna Gurevych, and Preslav Nakov. 2024.
\newblock \href {https://arxiv.org/abs/2404.14183} {Semeval-2024 task 8: Multidomain, multimodel and multilingual machine-generated text detection}.
\newblock \emph{Preprint}, arXiv:2404.14183.

\bibitem[{Wang et~al.(2025)Wang, Shelmanov, Mansurov, Tsvigun, Mikhailov, Xing, Xie, Geng, Puccetti, Artemova, Su, Ta, Abassy, Elozeiri, Ahmed, Goloburda, Mahmoud, Tomar, Aziz, Laiyk, Afzal, Koike, Kaneko, Aji, Habash, Gurevych, and Nakov}]{wang2025genai}
Yuxia Wang, Artem Shelmanov, Jonibek Mansurov, Akim Tsvigun, Vladislav Mikhailov, Rui Xing, Zhuohan Xie, Jiahui Geng, Giovanni Puccetti, Ekaterina Artemova, Jinyan Su, Minh~Ngoc Ta, Mervat Abassy, Kareem Elozeiri, Saad El~Dine Ahmed, Maiya Goloburda, Tarek Mahmoud, Raj~Vardhan Tomar, Alexander Aziz, Nurkhan Laiyk, Osama~Mohammed Afzal, Ryuto Koike, Masahiro Kaneko, Alham~Fikri Aji, Nizar Habash, Iryna Gurevych, and Preslav Nakov. 2025.
\newblock {GenAI} content detection task 1: {E}nglish and multilingual machine-generated text detection: {AI} vs. human.
\newblock In \emph{Proceedings of the 1st Workshop on GenAI Content Detection (GenAIDetect)}, Abu Dhabi, UAE. International Conference on Computational Linguistics.

\bibitem[{Zellers et~al.(2019)Zellers, Holtzman, Yu et~al.}]{zellers2019defending}
Rowan Zellers, Aidan Holtzman, Anna Yu, et~al. 2019.
\newblock Defending against neural fake news.
\newblock In \emph{Proceedings of NeurIPS 2019}, pages 10089--10101.

\end{thebibliography}

\appendix

\section{Appendix}
\label{sec:appendix}

\begin{table}[ht]
    \centering
    \small
    \renewcommand{\arraystretch}{1.25} 
    \setlength{\tabcolsep}{2.5pt} 
    \begin{tabular}{p{4cm}|>{\centering\arraybackslash}p{3cm}}
        \hline \hline
        \textbf{Tools \& Libraries} & \textbf{Version} \\
        \hline
        Python & 3.10.14 \\
        Pandas & 2.2.2 \\
        NumPy & 1.26.4 \\
        PyTorch & 2.4.0 \\
        Transformers & 4.44.2 \\
        Evaluate & 0.4.3 \\
        WandB & 0.16.6 \\
        \hline
    \end{tabular}
    \caption{Main tools and libraries used in our system}
    \label{tab:library_versions}
\end{table}

Table \ref{tab:library_versions} provide the details about the
corresponding libraries, which are beneficial to help replicate
our experiments.

\end{document}